\documentclass[10pt, a4paper]{article}
\usepackage{lrec}
\usepackage{graphicx}
\usepackage{tabularx}
\usepackage{soul}
\usepackage{comment}
\usepackage{epstopdf}
\usepackage[latin1]{inputenc}

\usepackage[draft]{hyperref}
\usepackage{xstring}
\usepackage{times}
\usepackage{latexsym}
\usepackage{url}
\usepackage{multirow}
\usepackage{threeparttable}
 \usepackage{todonotes}

\title{The Discussion Tracker Corpus of Collaborative Argumentation}

\name{Christopher Olshefski, Luca Lugini, Ravneet Singh, Diane Litman, Amanda Godley}

\address{University of Pittsburgh \\
         \{cao48, lul32, ras306, dlitman, agodley\}@pitt.edu\\}

\abstract{Although Natural Language Processing (NLP) research on argument mining has advanced considerably in recent years, most studies draw on corpora of asynchronous and written texts, often produced by individuals. Few published corpora of synchronous, multi-party argumentation are available. The Discussion Tracker corpus, collected in American high school English classes, is an annotated dataset of transcripts of spoken, multi-party argumentation. The corpus consists of 29 multi-party discussions of English literature transcribed from 985 minutes of audio. The transcripts were annotated for three dimensions of collaborative argumentation: argument moves (claims, evidence, and explanations), specificity (low, medium, high) and collaboration (e.g., extensions of and disagreements about others' ideas). In addition to providing descriptive statistics on the corpus,  we provide performance benchmarks and associated code for predicting each dimension separately,   illustrate the use of the multiple annotations in the corpus to improve performance via multi-task learning, and finally discuss other ways the corpus might be used to further NLP research.\\*\\*
}

\begin{document}

\maketitleabstract

\section{Introduction}

Natural Language Processing (NLP) research has advanced considerably in recent years, developing reliable predictors for argumentation \cite{mirkin2018listening,lippi2016argument,Aharoni:14,Biran:11,Carlile:18,Habernal:14,Park:18,Stab:14,Stab:17}, specificity \cite{Li:15,Li:16,Louis:12,Gao:19}, and collaboration \cite{richey2016sri}. The majority of corpora informing these developments are made up of written texts (e.g., documents, asynchronous online discussions) extracted from the web and often written by individuals. Additionally, annotations in these corpora typically focus on one linguistic dimension at a time (e.g., specificity \textit{or} dimensions of argumentation). As such, few published corpora both (1) focus on synchronous multi-party argumentation and (2) include multiple simultaneous annotations.

To address the lack of multi-party synchronous argumentation corpora that include multiple simultaneous annotations, we are releasing the Discussion Tracker corpus, which includes transcriptions of 29 multi-party synchronous argument-based dialogues collected in high school English classes and annotated for three simultaneous but distinct discourse dimensions (argumentation, specificity and collaboration). In addition to providing descriptive statistics on the corpus, we provide code and benchmark performance figures for predicting each of the three annotated dimensions, illustrating the challenging nature of this corpus.  We then illustrate the benefits the corpus has already afforded NLP algorithms for improving argument mining with multi-task learning and discuss other potential uses of the corpus for further NLP research.



\section{Related Work}

The development of argument mining tasks has been heavily informed by corpora of written texts extracted mostly from the web (e.g., online newspapers, Wikipedia, blog posts, user comments) \cite{al2018modeling,Brian:11,Aharoni:14,Habernal:14,Park:18,Swanson:15,Cabrio:14,Boltuzic:14}, from student essays \cite{Stab:14,Stab:17,Carlile:18}, from legal documents (parliamentary records and court reports) \cite{Mochales:11,Ashley:13}, or from speeches \cite{mirkin2018listening,lippi2016argument}. Similarly, NLP developments in specificity have typically drawn on newspaper articles \cite{Louis:11,Li:15,Li:16} or online content \cite{Gao:19,Ko:19}. Although work in Computer Supported Collaborative Learning (CSCL)  \cite{Weinberger:06,Fischer:13,Noroozi:13,Scheuer:14,Dillenbourg:08} has used mainly multi-party online data, it has typically drawn from asynchronous written discourse as opposed to synchronous dialogue. 

Many corpora of written text data that have been annotated for similar dimensions as the Discussion Tracker corpus have focused on single and independent tasks like argument components (e.g. claims, premises) \cite{Brian:11,Aharoni:14,Habernal:14,Mochales:11}, relations between pairs of arguments (e.g. support, attack) \cite{Park:18}, or specificity \cite{Li:15,Li:16,Louis:12}. Some corpora include multi-level annotations that allow for performing multiple tasks (e.g. argument components and relations) \cite{al2018modeling,Stab:14,Stab:17,Ashley:13,Carlile:18}. However, multi-level annotations like these have often been heavily dependent on one another. For example, although Carlile et al. analyzed two simultaneous annotations (argument components and specificity), their definition of specificity was contingent on argument component type, defining specificity of claims differently from specificity of premises. In this way, their multi-level annotations were tightly coupled and could only be analyzed in conjunction with one another. In cases like Al Khatib et al. \shortcite{al2018modeling}, which provides distinct multi-level annotations for asynchronous on-line argumentation, they do not include specificity at all. By providing annotations for multi-level and distinct annotations for collaboration, argumentation, and specificity in synchronous dialogues in natural learning environments, we believe the Discussion Tracker corpus is capable of offering the NLP community new opportunities for research.

Of the extant multi-party synchronous data, the most similar corpus to ours  is Richey et al.'s SRI corpus \cite{richey2016sri}. Their corpus was developed for analysis of the ways student talk patterns correlated with collaborative learning. Equipped with multi-speaker, small group audio recordings of middle school students discussing mathematical solutions, the SRI corpus provides multi-level annotations of collaboration indicators and collaboration quality. However, although the SRI corpus includes multi-party synchronous argumentation data, some major differences stand out from the Discussion Tracker corpus. First, the scope of the the Discussion Tracker corpus extends beyond collaborative dimensions to include dimensions of argumentation and specificity as well, whereas the SRI corpus focuses only on collaboration. Second, whereas the group size of the multi-party dialogues in the SRI corpus is three students, the group sizes in the Discussion Tracker corpus average around 15 students per discussion. Third, in addition to providing the gender of speakers, as the SRI corpus does, the Discussion Tracker corpus also includes identification of the racial background of speakers. Fourth, Discussion Tracker will be released as written transcriptions with corresponding annotations whereas the SRI data is released as audio files with time-stamped annotations. The differences between these corpora are not to suggest one is better than the other, but simply that different formats afford different avenues for analysis. 

\begin{table*}[!htb]
\centering
\begin{tabular}{|l|l|r|l|r|l|r|r|r|r|r|}
\hline
Tchr & M/F & \begin{tabular}[c]{@{}l@{}}Tchr\\  Exp\end{tabular} & Loc. & Grade & Course & \begin{tabular}[c]{@{}l@{}}Class\\ size\end{tabular} & Male & Fem. & \begin{tabular}[c]{@{}l@{}}Race\\ White\end{tabular} & \begin{tabular}[c]{@{}l@{}}Race\\ Non-white\end{tabular} \\ \hline
1 & F & 12 & suburb & 10 & regular & 29 & 20 & 9 & 28 & 1 \\ \hline
2 & M & 18 & suburb & 11 & honors & 11 & 5 & 6 & 7 & 4 \\ \hline
3 & F & 6 & suburb & 9 & honors & 16 & 11 & 5 & 8 & 8 \\ \hline
4 & M & 12 & suburb & 12 & AP & 13 & 4 & 9 & 6 & 7 \\ \hline
5 & F & 15 & urban & 9 & regular & 16 & 11 & 5 & 13 & 3 \\ \hline
6 & F & 14 & urban & 11 & AP & 18 & 11 & 7 & 14 & 4 \\ \hline
7 & F & 30 & urban & 10 & regular & 25 & 15 & 10 & 21 & 4 \\ \hline
8 & F & 30 & rural & 12 & AP & 13 & 3 & 10 & 12 & 1 \\ \hline
9 & F & 20 & rural & 10 & regular & 15 & 11 & 4 & 15 & 0 \\ \hline
10 & M & 20 & rural & 9 & honors & 25 & 16 & 9 & 25 & 0 \\ \hline
\end{tabular}
\caption{Characteristics of 10 classroom environments. Columns from left to right: teacher, teacher's gender (male or female), years of teaching experience, school location, grade level, course type, class size, number of male students, female students, white students, non-white students.}
\label{tab:Classrooms}
\end{table*}

\section{Data Collection}
\begin{table}[!htb]
\begin{tabular}{|l|l|r|r|r|}
\hline
Disc.  & Text & Stu. & Min. & \#Trns \\ \hline \hline
1a & \begin{tabular}[c]{@{}l@{}}Death of \\ Ivan Illych\end{tabular} & 27 & 40 & 208 \\ \hline
1b & Night & 28 & 41 & 134 \\ \hline
1c & The Name & 24 & 42 & 216 \\ \hline \hline
2a & \begin{tabular}[c]{@{}l@{}}Lgnd of \\ Slpy Hllw\end{tabular} & 11 & 32 & 49 \\ \hline
2b & \begin{tabular}[c]{@{}l@{}}The Mnstr's \\ Black Veil\end{tabular} & 9 & 19 & 43 \\ \hline
2c & \begin{tabular}[c]{@{}l@{}}Dickinson \\ Poems\end{tabular} & 10 & 33 & 110 \\ \hline \hline
3a & \begin{tabular}[c]{@{}l@{}}Lord of \\ the Flies\end{tabular} & 16 & 40 & 99 \\ \hline
3b & \begin{tabular}[c]{@{}l@{}}To Kill \\ A Mbird\end{tabular} & 12 & 35 & 81 \\ \hline
3c & \begin{tabular}[c]{@{}l@{}}To Kill \\ A Mbird\end{tabular} & 14 & 41 & 77 \\ \hline \hline
4a & \begin{tabular}[c]{@{}l@{}}Heart of \\ Darkness\end{tabular} & 13 & 44 & 109 \\ \hline
4b & \begin{tabular}[c]{@{}l@{}}Scarlett Ltr\end{tabular} & 13 & 45 & 63 \\ \hline
4c & \begin{tabular}[c]{@{}l@{}}A Mdsmmr \\ Night's  Drm\end{tabular} & 11 & 39 & 119 \\ \hline \hline
5a & \begin{tabular}[c]{@{}l@{}}To Kill \\ A Mbird\end{tabular} & 15 & 29 & 106 \\ \hline
5b & \begin{tabular}[c]{@{}l@{}}Smthg Wckd \\ This Wy Cms\end{tabular} & 16 & 35 & 105 \\ \hline
5c & \begin{tabular}[c]{@{}l@{}}The Little \\ Prince\end{tabular} & 15 & 38 & 111 \\ \hline \hline
6a & \begin{tabular}[c]{@{}l@{}}The Immortal \\ Lf of H. Lcks\end{tabular} & 6 & 35 & 124 \\ \hline
6b & The Crucible & 6 & 38 & 79 \\ \hline
6c & Into the Wild & 7 & 33 & 293 \\ \hline \hline
7a & \begin{tabular}[c]{@{}l@{}}Of Mice \\and Men\end{tabular} & 25 & 25 & 192 \\ \hline
7b & \begin{tabular}[c]{@{}l@{}} Fahr. \\ 451\end{tabular} & 11 & 34 & 332 \\ \hline
7c & MLK Jr. & 13 & 38 & 141 \\ \hline \hline
8b & \begin{tabular}[c]{@{}l@{}}The Yellow \\ Wllppr\end{tabular} & 11 & 39 & 127 \\ \hline
8c & Antigone & 13 & 28 & 248 \\ \hline \hline
9a & \begin{tabular}[c]{@{}l@{}}Salem Witch \\ Trials\end{tabular} & 15 & 35 & 275 \\ \hline
9b & The Crucible & 14 & 25 & 264 \\ \hline
9c & \begin{tabular}[c]{@{}l@{}}The Prks of \\ Bng Wllflwer\end{tabular} & 15 & 38 & 267 \\ \hline \hline
10a & \begin{tabular}[c]{@{}l@{}}Bleachers 
\end{tabular} & 23 & 38 & 165 \\ \hline
10b & JFK Speech & 25 & 33 & 148 \\ \hline
10c & \begin{tabular}[c]{@{}l@{}}To Kill \\ A Mbird\end{tabular} & 25 & 33 & 188 \\ \hline
\end{tabular}
\caption{Overview of discussions by teacher. ``Disc'' = Discussion, numbers correspond to teacher in Table \ref{tab:Classrooms}; ``Text''=the titles of the texts under discussion; ``Stu.''= amount of students present in discussion; ``Min.'' = length in minutes; ``\#trns''= number of turns per discussion}
\label{tab:discussions}
\end{table}
The Discussion Tracker corpus is based on audio-recorded multi-party spoken discussions in 10 different high school English teachers' classrooms across three different school districts (suburban, urban, and rural) (see Table \ref{tab:Classrooms}). Between October 2018 - March 2019 we recorded a total of three literature discussions per classroom. Omitting one discussion that was off-topic, the corpus we are releasing contains 29 transcriptions of high school literature discussions based on 985 minutes of audio (see Table \ref{tab:discussions}).  

Across the ten classrooms, the mean number of discussion participants was 15 students (SD 6), ranging from 6 to 29.
In accordance with educational research examining racial and gender inequities in instructional practice \cite{kelly2008race,sherry2014indirect}, we collected metadata for race and gender demographics. Based on notes taken during data collection, we estimated that on average student discussants were 50\% male and 50\% female (SD 0.18), 77\% white (SD 0.22) and 23\% nonwhite (SD 0.22). Of the nonwhite students most appeared to be Indian (58\%), 18\% appeared Black, 15\% East Asian, and the remaining 9\% appeared Latinx or other. 


In order to maintain the authenticity of the instructional environment, teachers were free to facilitate their discussions according to their pedagogical expertise, so long as they arranged students to face the microphone (which was placed in the center of the classroom) and attempted to ensure that students sat in the same seats for each discussion. Speaker demographics varied slightly across discussions within the same classroom due to student absences and discussion styles (e.g., holding a small group discussion in which only a portion of the students were expected to speak). Descriptions of the classrooms (Table \ref{tab:Classrooms}) reflect the maximum number of possible discussants when all students were present. Any variation between these classroom descriptions and the descriptions of the discussions within each classroom (Table \ref{tab:discussions})  can be explained either by student absences or discussion style.

In most discussions, the entire class participated, although five discussions were set up as small groups and thus limited to a subset of the students present (see 6a, 6b, 6c, 7b, 7c in Table \ref{tab:discussions}). Discussions were recorded by a member of the research team using a Zoom H6 six-track portable audiorecorder placed in the center of the classroom with student discussants arranged in circles around the device. In addition to creating a map that linked numerical IDs to the locations of each discussant, a researcher also kept handwritten notes to identify speakers. Transcriptions of the audio data were outsourced to a professional service (Rev.com) and were reviewed by research assistants for accuracy. In a test of four transcripts, an average of 4\% words per transcript were incorrectly transcribed and required revision. In addition to aligning speaker IDs to transcribed talk, research assistants also transferred data to an excel document formatted specifically for annotation. 

\section{Data Annotation}
The Discussion Tracker corpus includes annotations for three dimensions of student talk that researchers in classroom discourse have associated with positive educational outcomes \cite{howe2019teacher,applebee2003,chisholm2011,soter2008,sohmer2009guided,juzwik2013,nystrand1991} (see Table \ref{fig:transcript}); similar dimensions have also been annotated by NLP researchers for other types of data. Using a classroom discussion annotation scheme optimized for NLP development \cite{lugini2018annotating}, we annotated student talk for {\it argument moves} (claims, evidence and explanations) and {\it specificity} (low, medium and high). In addition, we developed an annotation scheme for {\it collaboration} that  synthesized findings in both classroom discourse research \cite{engle2002guiding,keefer2000} as well as the computer-supported collaborative learning (CSCL) literature \cite{samei,zhangcscl}.

Prior to annotation, speakers were identified using the handwritten notes taken during data collection. Cases in which speaker IDs were difficult to determine were labeled either as `St?' if the speaker was likely a student or `?' if it was unclear whether the speaker was a student or teacher. Student talk was segmented into both turns at talk and argument discourse units. Talk at the turn level was annotated for collaboration, and talk at the argument discourse unit was annotated for argument move values (claim, evidence, explanation) and specificity level (low, medium, high). The coding instructions for all  annotated dimensions are briefly reviewed below. More details can be found in Lugini et al. \shortcite{lugini2018annotating}, and a sample coded transcript with all annotation manuals can be found in the link provided in section \ref{sec:public}    

Similar to argument mining systems like Nguyen \shortcite{nguyen2018argument}, our pipeline for annotating collaborative argumentation involved several steps. (1) While examining only student talk, we flagged turns  that contained no substantive argumentation and were thereby deemed \textit{non-argumentative}. Turns that included both non-argumentative and argumentative phrases were considered argumentative. In addition to talk that was inaudible or off-topic (``I have to go to the bathroom''), non-argumentative talk also included meta-discourse talk (``okay you can take a turn,''), discussion prompts (``Describe the imagery in the poem'') and brief agreements (``yeah'').
(2) Argumentative turns  were annotated for one of four collaboration dimensions, \textit{New Ideas}, \textit{Agreements}, \textit{Extensions}, and \textit{Probes/Challenges}. (3) Turns containing collaborative argumentation were further segmented into argument discourse units, which were, (4) labeled for argument move type (\textit{claims}, \textit{evidence}, or \textit{explanations}), and (5) annotated for specificity. 

\subsection{Collaboration}

Each argumentative turn  was annotated for one of four possible collaborative relationships with prior turns at talk. (1) \textit{New Ideas}: turns that did not reference ideas in prior turns at talk. (2) \textit{Agreements}: turns that repeated verbatim or almost verbatim the idea in a prior turn. (3) \textit{Extensions}: turns that built on prior ideas, either the speaker's own or another student's. (4) \textit{Probes/Challenges}:  turns that questioned or disagreed with a prior idea. Also included in collaboration annotations was a reference to the prior turn with which the current turn was in a collaborative relationship (turn reference number). After coding approximately one-third of the transcripts, analyses revealed that  30\% of turns had a collaborative relationship with one of the previous two turns, and 95\% had a collaborative relationship with turns within the previous four turns. Thus a limit for turn references was set at no more than four annotated previous turns unless the speaker's reference to an earlier turn was explicit (e.g., a speaker said, ``Going back to John's comment about authority" when John had commented 10 turns previously).    

\subsection{Segmentation}
Prior to annotating for argumentation and specificity, argumentative turns at talk were segmented further into argument discourse units (ADUs). Similar to Ghosh et al. \shortcite{ghosh2014analyzing}, who segmented ADUs into either ``stance'' vs ``rationale,'' annotators were instructed to divide turns at talk into interpretive vs. factual/ information-based segments of talk. For example, as seen in Table \ref{fig:transcript}, Speaker 1's turn at talk was first segmented when they offer examples ``throughout history'' of their claim. The turn was segmented a second time when the speaker offered an interpretation of how the examples related to their claim. Annotators were not expected to get so fine-grained as Stab and Gurevych \shortcite{stab2017parsing}, whose ADU segmentation accounted for sub-claims and multiple premises. Thus, annotators were instructed not to segment turns into multiple claims or multiple units of evidence.

\subsection{Argumentation}

As in Lugini et al.~\shortcite{lugini2018annotating}, annotations for argumentation were derived from classical models of argument structure \cite{toulmin1958}, and were simplified to include three labels: \textit{claim} (an arguable statement that presents an interpretation of a text or topic), \textit{evidence} (facts or information to support a claim) and \textit{explanation} (reasoning or justification for why the given evidence supports the claim).

\subsection{Specificity}
Our annotation scheme for specificity differs from Carlile et al. \shortcite{Carlile:18} in that our  labels were not contingent on argumentation, but rather stood independent of argumentation labels. Like Lugini et al. \shortcite{lugini2018annotating}, we defined specificity as the existence of particularity, detail, content-language (use of disciplinary terminology like ``symbolism'' or ``irony''), and/or a chain of reasons. Argument units that included two or more of these four characteristics above were annotated as {\it high specificity}; argument units containing one of the characteristics were annotated as {\it medium specificity}; and argument units containing none of the above characteristics were annotated as {\it low specificity}. 
\begin{table*}[]
\begin{tabular}{|c|c|p{2.5in}|c|c|c|c|}
\hline
\textbf{Turn}      & \textbf{Speaker}       & \textbf{Talk}                                                                                                                                                                                                                                                                                                                                                                                                                                                    & \textbf{\begin{tabular}[c]{@{}c@{}}Collab-\\ oration\end{tabular}} & \textbf{\begin{tabular}[c]{@{}c@{}}Reference\\ Turn\end{tabular}} & \textbf{\begin{tabular}[c]{@{}c@{}}Argument-\\ ation\end{tabular}} & \textbf{\begin{tabular}[c]{@{}c@{}}Spec-\\ ificity\end{tabular}} \\ \hline
\multirow{3}{*}{1} & \multirow{3}{*}{St 1}  & My interpretation of it is that, without a middle ground, you are left with two very extreme points. Whether or not the middle ground directly centered, we have a range. We have a spectrum.[...]                       & \multirow{3}{*}{New}                                               & \multirow{3}{*}{}  & Claim                                                              & Medium                                                           \\ \cline{3-3} \cline{6-7} 
                   &                        & Throughout history, whether you go back to ancient Europe, and you look at tyrannies and dictatorships, not even ancient Europe. If you go back to the Holocaust and what Hitler was doing over in Germany [...] if you go back to Communism, as well [...]                                                                                                                                                                                                  &                                                                    &                    & Evidence                                                           & Medium                                                           \\ \cline{3-3} \cline{6-7} 
                   &                        & Those are two extremes, and neither of them ended well, and just anarchy there. There is no order there, there is no civilized kind of society to base anything around. I think the middle ground is necessary just to create some kind of spectrum that we can go off of.                                                                                                                                                                                       &                                                                    &                    & Explanation                                                        & Medium                                                           \\ \hline
2                  & St 9                   & I acknowledge your point, but there wasn't nobody going against anything until this happened, until this event occurred.                                                                                                                                                                                                                                                                                                                                         & Challenge                                                          & 1                  & Claim                                                              & Low                                                              \\ \hline
3                  & St 1                   & Does that make the way they were living right, thought?                                                                                                                                                                                                                                                                                                                                                                                                          & Challenge                                                          & 2                  & Claim                                                              & Low                                                              \\ \hline
4                  & St 9                   & If they were happy, I believe they were perfectly fine.                                                                                                                                                                                                                                                                                                                                                                                                          & New                                                                &                    & Claim                                                              & Low                                                              \\ \hline
5                  & St 17                  & My assessment of the topic at hand is, there needs to be a balance between state rights and user rights. [xx] slide, and to what extent was it off balance.                                                                                                                                                                                                                                                                                                  & Extension                                                          & 1                  & Claim                                                              & Medium                                                           \\ \hline
6                  & St 14                  & I concur with both St 19 and 17's statements. I also think that if we have society in which people are afraid to go against the core, then the rights of them are restricted. They're afraid that if they step outside the lines, then it won't end good for them, so everybody's afraid [...] they won't be accepted.                                                                                                                                         & Extension                                                          & 5                  & Claim                                                              & Medium                                                           \\ \hline
\multirow{2}{*}{7} & \multirow{2}{*}{St 18} & I concur with St 14.                                                                                                                                                                                                                                                                                                                                                                                                                                             & \multirow{2}{*}{Extension}                                         & \multirow{2}{*}{6} & Claim                                                              & Low                                                              \\ \cline{3-3} \cline{6-7} 
                   &                        & Because back in the day when we have the Civil War going on, people were on different sides. People were afraid to come and say, ``oh, I'm in between,'' because then they would be afraid that they'd be treated just like African Americans. As St.,9 said, it got to the point where they were on two different sides and they couldn't decide on something, so they said, ``hey, let's fight this out, and whoever wins basically decides what action happens.'' &                                                                    &                    & Evidence                                                           & High                                                             \\ \hline
\end{tabular}
\caption{Sample transcript from discussion 6b on the play ``The Crucible''.}
\label{fig:transcript}
\end{table*}

\subsection{Reliability Analyses}
Reliability analyses of segmentation and annotations yielded high inter-rater agreement. We calculated reliability measures for each of the following categories:  
1) selecting collaborative argumentative turns at talk, 2) annotations for collaboration labels, 3) segmentation of turns into argument discourse units, 4) annotations for each argument move label and 5) specificity labels. 

Recall that annotators were instructed to consider turns at talk as \textit{non-argumentative} if they did not include argumentation. 
Agreement between annotators for argumentative turns versus not was based on a sample with highly skewed class distributions yielding low Kappa (0) but high raw percentage of agreement (92\%). 

Basing our argumentative turns segmentation metric on Habernal and Gurevych \shortcite{Habernal:17}, we computed agreement on 54 transcripts from a classroom corpus we collected before Discussion Tracker by comparing the segmentation of two trained annotators. The average alpha was 0.96, min was 0.72, max was 1, and standard deviation was 0.048. All data in the Discussion Tracker corpus was segmented by the same trained annotator. 

A portion of the transcripts were double-annotated and yielded substantial agreement for collaboration (Cohen's Kappa, 0.74) and specificity (Quadratic Weighted Kappa, .70), and near-perfect agreement for argumentation (Cohen's Kappa, .89). Because our specificity annotation was based on an ordered scale (low, medium, high), we employed an inter-rater agreement measure (Quadratic weighted Kappa) that could account for degrees of disagreement. Because argument move annotations were not based on an ordered scale we simply used Cohen's Kappa in which disagreements were not specially weighted. 

After achieving satisfactory agreement in double-coding, the remaining transcripts were single-annotated for collaboration, argumentation, and specificity. 
Differences between annotators were resolved through deliberation  to construct gold standard annotations for public release.

\subsection{Corpus Statistics}

Of Discussion Tracker's 3261 student turns, 2128 were considered \textit{argumentative} and were annotated for collaborative argumentation. As seen in Table \ref{tab:stats}, the large majority of annotated turns were labeled as either \textit{New Ideas} (37.69\%) or \textit{Extensions} (47.70\%), with \textit{Challenges/Probes} and \textit{Agreements} making up only a narrow portion of the corpus. 

\begin{table}[!t]
\begin{tabular}{|l|l|r|r|}
\hline
 & Annotation & Total Count & Percentage \\ \hline
\textbf{\begin{tabular}[c]{@{}c@{}}Collab-\\ oration\end{tabular}} & New & 802 & 37.69\% \\ \cline{2-4} 
 & Agree & 37 & 1.74\% \\ \cline{2-4} 
 & Extensions & 1015 & 47.70\% \\ \cline{2-4} 
 & Chall/ Probes & 274 & 12.88\% \\ \cline{2-4} 
 & \textbf{Total} & \textbf{2128} & \textbf{100.00\%} \\ \hline
\textbf{\begin{tabular}[c]{@{}l@{}}Argum-\\ entation\end{tabular}} & Claims & 2047 & 65.30\% \\ \cline{2-4} 
 & Evidence & 762 & 24.31\% \\ \cline{2-4} 
 & Explanations & 326 & 10.40\% \\ \cline{2-4} 
 & \textbf{Total} & \textbf{3135} & \textbf{100.00\%} \\ \hline
\textbf{Specificity} & Low & 1189 & 37.93\% \\ \cline{2-4} 
 & Medium & 1071 & 34.16\% \\ \cline{2-4} 
 & High & 875 & 27.91\% \\ \cline{2-4} 
 & \textbf{Total} & \textbf{3135} & \textbf{100.00\%} \\ \hline
\end{tabular}
\caption{Descriptive statistics of corpus annotations.}
\label{tab:stats}
\end{table}

The argumentative turns at talk were further segmented into 3135 argumentative discourse units to be annotated for argument move type and specificity. The corpus is made up of mostly claims (65.30\%),  less than half of which were supported with evidence (24.31\%), and still less were elaborated with explanations (10.40\%). Specificity of argument moves was more evenly distributed with 37.93\% annotated as low, 34.16\% as medium, and 27.91\% high. 

\section{Public Release}
\label{sec:public}
The Discussion Tracker Corpus will be freely available for research purposes, with the release coordinated with the publication of this paper. The release will include 29 separate .xlsx documents segmented for both turns at talk and argument discourse units. Each document will contain the discussion in full, including teacher talk and non-annotated student talk for context. Additionally, transcripts will contain unique ID numbers (e.g., T127.1.Heartdark) for each turn at talk indicating the de-identified teacher (e.g., T127), 
the discussion (e.g., .1-- referring to the first discussion from that teacher's classroom), the text (e.g., ``Heartdark'' for Joseph Conrad's \textit{Heart of Darkness} and the turn number in the discussion (the final number in the ID). Directly to the right of each turn at talk will be columns containing collaboration annotations and their corresponding turn reference numbers. Talk segmented at the ADU level will include annotations for each argument move type and specificity. 

The corpus is available at \url{https://discussiontracker.cs.pitt.edu}. Included in the corpus are 29 transcripts with complete annotations, a metadata file containing information for the speaker demographics and grade levels, and the coding manuals for creating the annotations.

The code for all classification experiments from Section~\ref{sec:case} is also available via the Discussion Tracker website to provide a performance benchmark for future  research that uses this corpus.

\section{Case Studies Using the Corpus}
\label{sec:case}
Our corpus provides NLP researchers  the opportunity of several uses.
First, each of the three annotated dimensions of collaborative argumentation can be used individually to train a classifier for automated prediction.  
Second, we believe that one of the most interesting characteristics of the Discussion Tracker corpus is the fact that it provides annotations for multiple dimensions of collaborative argumentation simultaneously.
It is possible, then, to analyze if and how these dimensions are related. If that is the case, it may be possible to develop more robust and accurate models for automated classification of such dimensions.
Carlile et al. \shortcite{Carlile:18} annotated argumentative discourse units for argument component and specificity (among other things) and were able to make use of these two aspects simultaneously to predict argument persuasiveness in written essays.
Similarly, we showed in our previous study that specificity can be used to improve the performance of argument component classifiers \cite{Lugini:18b}. 
We found that models trained through multi-task learning where the primary task consists of argument component classification and the secondary task consists of specificity classification almost always outperform models that only perform argument component classification.
However, the corpus used in our previous study is not publicly available and therefore our previous results are not reproducible by other members of the research community.

To provide reproducible performance baselines to facilitate future classifier evaluations using the Discussion Tracker corpus, we present our experiments in learning models for individual classification tasks and then jointly learning multiple classifiers.
The performance of each model was evaluated using the same ten-fold cross-validation: each fold consists of 26 transcripts as training set and 3 as test set (except for one fold where 27 transcripts are used for training and 2 for testing). We report accuracy, Cohen Kappa (quadratic-weighted for specificity and unweighted for the remaining tasks) and macro f-score as evaluation metrics.
The particular folds used for the cross-validation experiments presented in this paper will  be made  available in the corpus release in the form of a json file containing a list of all training and test transcripts for each fold.

\subsection{Learning through Individual Annotations}
\subsubsection{Argumentation}
\label{sec:case_study_argumentation}
As a baseline for evaluating the performance of argument component classification on the Discussion Tracker corpus, we tested our previously proposed model \cite{Lugini:18b}, which showed significantly higher performance on a previously examined set of classroom discussions compared to argument mining models developed for other types of corpora.
It consists of a hybrid model which combines embeddings generated through a neural network with a set of handcrafted features.

The handcrafted features consist primarily of two sets: Speciteller feature set, derived from prior work on specificity \cite{Li:15}, and online dialogue feature set, derived from prior work on argument mining in online dialogues \cite{Swanson:15}.
The Speciteller feature set contains the following features, extracted independently for each argument move: average of 100-dimensional word vectors \cite{Turian:10} for words in the argument move, number of connectives, number of words, number of symbols, number of numbers, number of capital letters, number of stopwords normalized by argument move length, average characters per word, number of subjective and polar words (extracted using the MPQA \cite{Wilson:09} and the General Inquirer \cite{Stone:63} dictionaries), average word familiarity (extracted using MRC Psycholinguistic Database \cite{Wilson:88}), and inverse document frequency statistics (maximum, minimum).
The online dialogue feature sets includes: number of pronouns, number of occurrences of words of different lengths, descriptive word-level statistics, term frequency - inverse document frequency of unigrams and bigrams (with frequency greater than 5), and part of speech tag features (unigrams, bigrams and trigrams).\footnote{One feature set from our prior work is not included in this study, namely wLDA \cite{Nguyen:16}, since it greatly increased model complexity leading to overfit.}

The neural network model is composed of a series of 3 convolutional/max-pooling layers, in which the convolutional layer consists of 16 filters of size 7.
Each word in an argument move is first processed through an embedding module which uses pretrained GloVe embeddings \cite{Pennington:14} of dimensionality 50.
The argument move is then processed through the neural network to generate a fixed-size embedding.
\begin{figure}[!thb]
\begin{center}
  \includegraphics[scale=0.37]{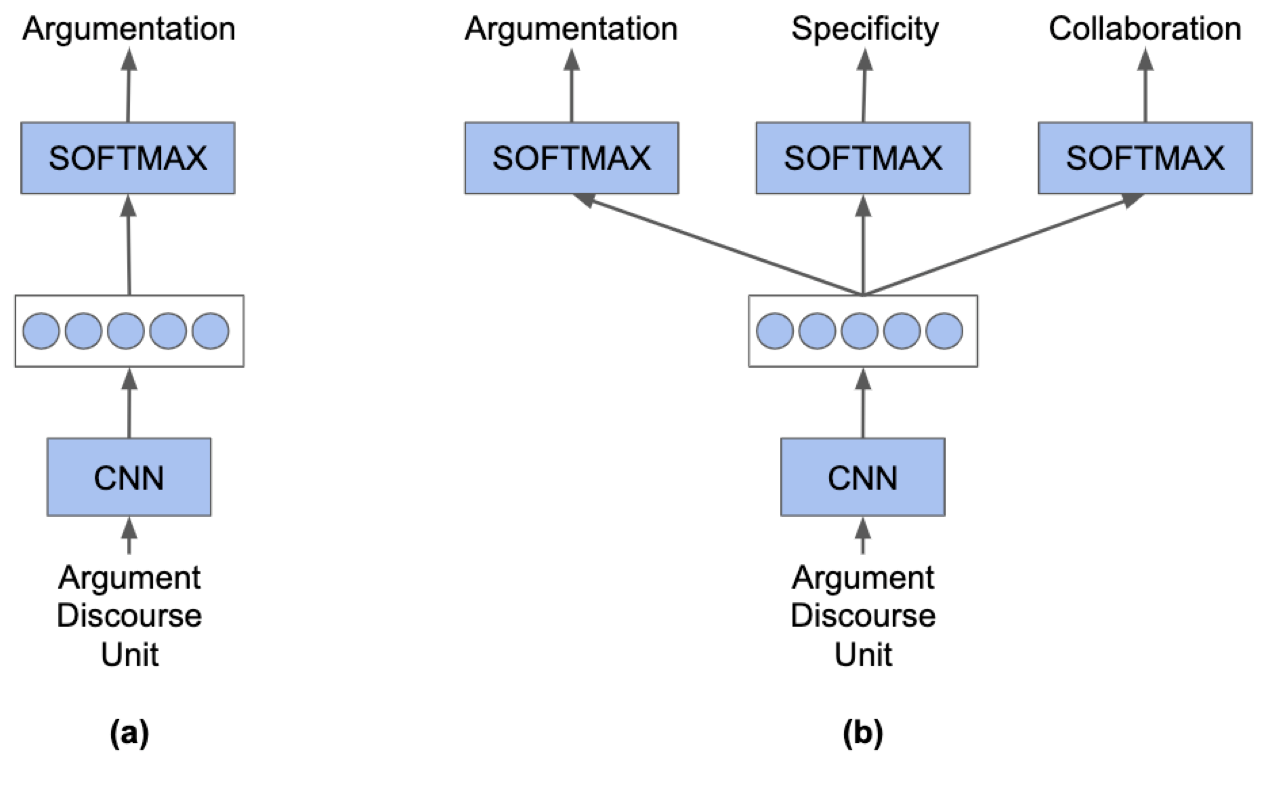}
  \end{center}
  \caption{Neural network models used in this study: individual models (a); joint multi-task model (b).}
\label{fig:neural_network_models}
\end{figure}
The handcrafted features are concatenated to the neural network embeddings, and the final feature vector is  input to a softmax classifier to output the final prediction (see Figure \ref{fig:neural_network_models}(a)).

In our prior work \cite{Lugini:18b} on argument component classification for discussions, we used oversampling to alleviate the class imbalance present in argumentation labels (which is also present in the Discussion Tracker corpus). However, since our Discussion Tracker experiments also include 3 task multi-task learning, oversampling with respect to argumentation labels might have negative impact on other tasks.
To address class imbalance, instead of using oversampling, we manually set class weights to impact the loss function, trying to increase the importance of the less frequent labels (evidence, explanation).
The class weights were set by roughly approximating the frequency of labels in the corpus used in our prior work \cite{Lugini:18b}:
the previous corpus contained almost twice as many claims than evidence, and almost twice as many explanations as evidence, prompting us to use the weights \textit{(claims: 1, evidence: 2, explanations: 4)}.

Table \ref{tab:results} shows the cross-validation results.
As we can see from Row 1, the difference between accuracy and f-score indicates that the model performs differently for the three argumentation labels.
The f-score for claims, evidence, and explanations are respectively 0.776, 0.565, and 0.164.
While specifying class weights at training time helped, this shows that there is ample room for improvement.

\subsubsection{Specificity}
As we showed in our previous study \cite{Lugini:17}, using the off-the-shelf Speciteller tool~\cite{Li:15} for predicting sentence specificity performed poorly when applied to text-based classroom discussions. We were able to significantly improve classification results by proposing features and models explicitly developed for classroom discussions.
Like our previous work on argumentation, however, the corpus is not publicly available.
To provide a  baseline for the Discussion Tracker corpus, we evaluated specificity prediction performance using the same model as described in Section \ref{sec:case_study_argumentation}
By using this model we achieved very similar performance to the model we proposed in Lugini and Litman \shortcite{Lugini:17} at a fraction of the computational cost. The main difference between the two models is the use of a convolutional neural network instead of a recurrent neural network.

As we can see in Row 3 of Table \ref{tab:results}, the small variation across the three performance metrics indicates consistent performance for all three specificity labels.
Additionally, kappa and f-score show that the specificity model is much more accurate than the one for argumentation.
\begin{table*}[!htb]
\centering
\begin{tabular}{cllccc}
\textbf{Row} & \textbf{Experiment} &Model & \textbf{Accuracy} & \textbf{Kappa} & \textbf{Macro F} \\ \hline
1 & Argument Move &Individual& 0.669 & 0.343 & 0.502 \\
2 & Argument Move &Joint & 0.673 & 0.365 & 0.516 \\ \hline
3 & Specificity &Individual& 0.703 & 0.750 & 0.695 \\
4 & Specificity &Joint & 0.706 & 0.751 & 0.698 \\ \hline
\end{tabular}
\caption{Neural classification results (both individual and joint models) for argument discourse unit prediction tasks.}
\label{tab:results}
\end{table*}

\subsubsection{Collaboration}
Since collaboration was not annotated in our previously used corpus of classroom discussions (unlike argumentation and specificity), 
we do not have a prior  prediction model to draw upon as a baseline. We instead use 
Naive Bayes to model collaboration, both for model simplicity and because it is a typical baseline model for NLP tasks. 
The feature vector we use is bag of words (BOW) on each turn, tokenized using NLTK's word tokenizer. We filter out  stop words and use tokens that occur once to approximate unknown words.
We experimented with using TF-IDF weighting on  bag of words 
and using different Naive Bayes variants.  Results from the best performing configurations  are reported in Table \ref{tab:collabResults}. 

As shown in Row 1, we found that the best configuration  for predicting the four collaboration labels  was Multinomial Naive Bayes with TF-IDF features. Note that  accuracy is much higher than both kappa and macro f-score, likely reflecting the skewed distribution among the four classes. These baseline results show the difficulty in distinguishing between all of the collaboration annotations in the Discussion Tracker corpus. 
We also explored a simpler binary version of our classification task (Row 2), which  reduced the class skew while still making a pedagogically useful distinction. 
In particular, during discussions with teachers where we visualized the collaboration annotations in the corpus that came from their particular classrooms, we found that teachers were very curious about whether students were introducing new information into the discussion or building off of what was previously said. Therefore we  experimented on how well a classifier could distinguish student turns labeled `New' from the other collaboration annotations. We found that Gaussian Naive Bayes without TF-IDF features performed the best. While the results improved compared to  predicting the original 4 classes, there is still  room for improvement. In sum, determining the  collaboration labels for  student turns  is difficult for our simple  Naive Bayes with BOW baseline method.


Finally, although the collaboration experiment was performed using only the manually-annotated argumentative subset of student turns (as is typical of system component evaluations), an end-to-end system would in addition need to first (or jointly) automatically separate the argumentative and non-argumentative turns, before classifying the collaboration labels for the argumentative turns. Using the same approach as for predicting collaboration labels, we find that a Gaussian Naive Bayes model with TF-IDF performs the best at this task (Row 3 in Table \ref{tab:collabResults}). 


\begin{table*}[bht]
    \centering
    \begin{tabular}{l l l r r r}
        \textbf{Row} & \textbf{Experiment} & \textbf{Model} & \textbf{Accuracy} & \textbf{Kappa} & \textbf{Macro F} \\ \hline
        1& Collaboration (all 4 labels) & Multinomial W/TF-IDF & 0.504 & 0.086 & 0.254 \\
        2& Collaboration (`New' vs Other)  & Gaussian w/ BOW & 0.623 & 0.217 & 0.604 \\ \hline
        3& Argumentative vs Non-Argumentative  & Gaussian W/TF-IDF & 0.785 &0.513 & 0.756 \\ \hline
        
    \end{tabular}
    \caption{Naive Bayes classification results for turn-level prediction tasks.}
    \label{tab:collabResults}
\end{table*}


\subsection{Learning through Multiple Annotations}
\label{sec:results}
In this section, we describe an experiment that extends our previous multi-task learning study \cite{Lugini:18b}, by using three rather than two tasks for the learning, by using the new Discussion Tracker corpus, 
and by providing annotation and associated benchmark results that can be replicated and extended by others in the future. More specifically, we test the following research hypotheses:
by modeling argument component, specificity, and collaboration information simultaneously, we can develop a single model that outperforms individual classifiers for (H1) argument component, and (H2) specificity.
\begin{figure}[!bt]
\begin{center}
  \includegraphics[scale=0.4]{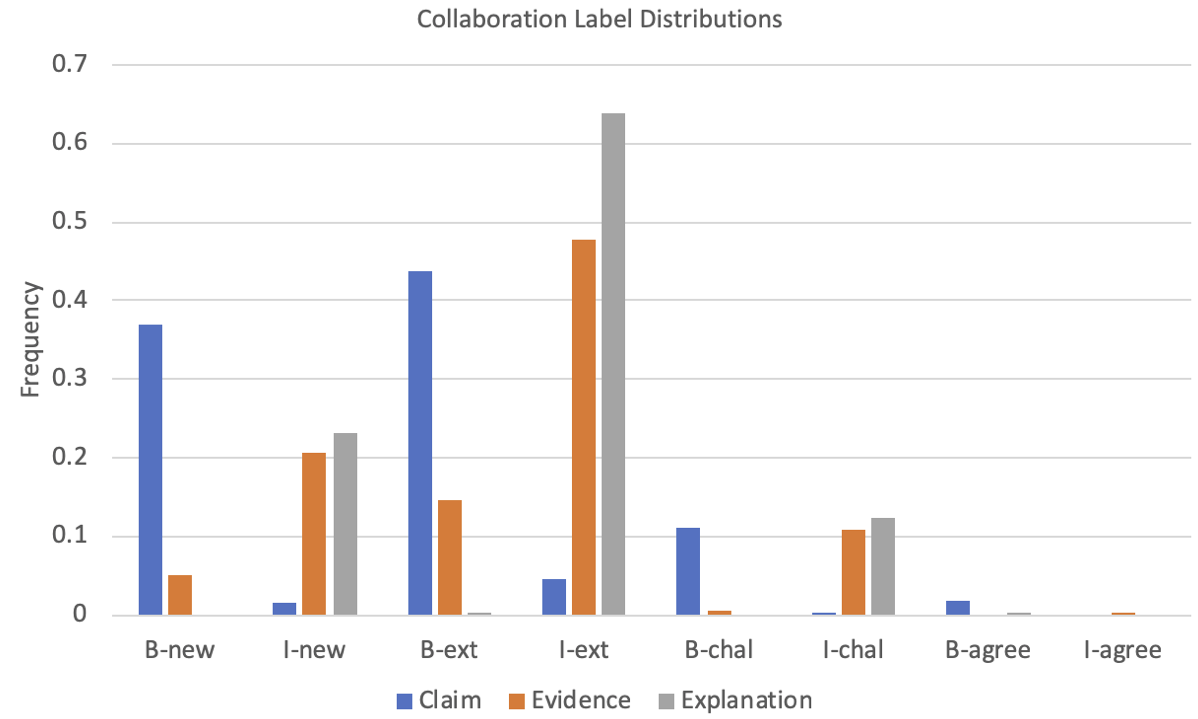}
  \end{center}
  \caption{Distribution of collaboration labels for different argumentation categories}
\label{fig:collaboration_label_distribution}
\end{figure}
These hypotheses are motivated by our observation of differences between collaboration label distributions across argumentative moves.
However, given the different unit of analysis for the annotation of collaboration (turn) versus argumentation and specificity (argument discourse unit), for the multi-task learning setting the collaboration annotations have been converted to BIO 
format in order to have one annotation per argument move\footnote{If a turn labeled ``\textit{New}'' for collaboration is segmented into two argument moves, the collaboration label is converted into ``\textit{B-New}'' for the first argument move and ``\textit{I-new}'' for the second.}.
For example, we observed that the most frequent collaboration labels for claims are \textit{B-extension $(43.9\%)$, B-new $(37.0\%)$ and B-challenge $(11.1\%)$}.
Looking at evidence, the most frequent collaboration labels are \textit{I-extension $(47.8\%)$, I-new $(20.8\%)$ and B-extension $(14.7\%)$}.
Lastly, the most frequent collaboration codes for explanations are \textit{I-extension $(63.9\%)$, I-new $(23.2\%)$ and I-challenge $(12.2\%)$}.
The complete label distributions are shown in Figure \ref{fig:collaboration_label_distribution}.

With the goal of exploiting the potential relationships between argumentation, specificity, and collaboration, we developed a single joint model trained through multi-task learning.
This model consists of the same convolutional neural network (see Section \ref{sec:case_study_argumentation}) along with the same handcrafted feature set, with the difference that the final feature vector is used as input to three softmax classifiers simultaneously: one for argument component, one for specificity and one for collaboration.
In this setting the representation for an argument move is entirely shared between the three tasks (see Figure \ref{fig:neural_network_models}(b)).
The final loss of the model is the sum of the individual cross-entropy losses: we chose an unweighted sum so that we can understand potential relationships between the three prediction tasks; if the goal is that of maximizing performance, one of the tasks can be favored by increasing its weight in the loss function.

Table \ref{tab:results} shows the results of our experiments.
Rows 1 and 2 relate to our hypothesis H1, and we can see an improvement in accuracy, kappa and f-score. The performance improvements achieved through the joint model, though, are not yet statistically significant.
Although differences exist across argumentation labels for different collaborative moves, our joint model is not able to optimally capture them. This may be due to the low performance of the collaboration classifier: if the collaboration model cannot reliably capture collaboration information, it cannot properly inform the argumentation classifier.
We believe that increasing the performance of the individual classifiers and using learned weights in the loss function  will result in a more effective joint model. Rows 3 and 4 relate to our hypothesis H2.
Like for argumentation, the results on specificity show the joint model outperforming the single-task one in all metrics, though the difference was not statistically significant.\footnote{Recall that while collaboration was  annotated at the turn level, the joint model uses the BIO converted (ADU) representation. We thus do not investigate whether the joint ADU model improves the individual turn-level collaboration prediction.} 


Although the current results show a limited gain in performance of the joint model over the individual ones, the Discussion Tracker corpus allows the research community to further analyze inter-dependencies between argumentation, specificity and collaboration, and develop more effective models to take advantage of these dependencies.

\subsection{Other Potential Corpus Uses}

Going beyond classification, our corpus can also be used in conjunction with other publicly available corpora.
Daxenberger et al. \shortcite{Daxenberger:17} for example performed a qualitative analysis to understand the difference in conceptualization of claims across multiple datasets. None of the datasets analyzed, however, includes transcripts of spoken dialogues.
The Discussion Tracker corpus can be used in a similar way, for example, to study the different conceptualization of argument components between spoken multi-party discussions and online multi-party dialogues or written essays.
Additionally, the corpus could also support educational research, which has taken interest in classroom discourse since the 1970's \cite{howe2013classroom,mercer2014study}. Howe et al.'s \shortcite{howe2019teacher} recent study of 72 elementary classroom environments established statistically significant relationships between positive learning outcomes and student talk dimensions similar to the annotations we include in our corpus: participation (how much and how many students speak), elaboration (similar to our extensions), and querying (similar to our challenge/probe category). 
The corpus metadata can also be used to support the investigation of issues of educational equity (e.g., gender and racial) in collaborative argumentation research \cite{godley2019,howe1997gender,kelly2008race}.


\section{Future Corpus Extensions}
Over the next three years of the Discussion Tracker project, we will be collecting and annotating new classroom data that will more than triple
the size of this first release of our corpus. In these future corpus extensions, we will also include new information in our transcripts, namely time stamps and  phenomena specific to spoken data (including filled pauses like ``uh''). This will allow for more investigation on the similarities and differences between spoken and written synchronous collaborative argumentation.\footnote{Although the inclusion of audio files would contribute greatly to these endeavors and might be possible in future releases, standard IRB regulations for research on minors in authentic learning environments would require costly de-identification.}

\section{Summary}
By releasing the Discussion Tracker corpus, we hope to contribute to collaborative argument mining research concerned with multi-party synchronous argumentative discourse collected in authentic environments. The 29 transcriptions included in the Discussion Tracker corpus contain multi-party argumentation along with annotations for collaboration at the turn level and annotations for argument move type and specificity at the argument discourse unit level. We created performance baselines for a variety of individual classification tasks, and demonstrated the potential use of the simultaneous annotations by exploring multi-task learning as method for improving  baseline performance. We believe that the Discussion Tracker corpus will be a useful resource for others, not only because it provides challenging multi-party collaborative argumentation data for future NLP research, but also because it provides multiple simultaneous annotations that can allow for a wider variety of learning approaches. 

\section{Acknowledgements}
This work was supported by the National Science Foundation (EAGER 1842334 and 1917673)
and in part by the University of Pittsburgh Center for Research Computing through the resources provided.

\section{Bibliographical References}
\label{main:ref}

\bibliographystyle{lrec}
\bibliography{main_text_LREC}


\end{document}